%% file: Subspace_learning.tex
\documentclass{article}

\usepackage[titletoc,title]{appendix}
\usepackage{times}
\usepackage{graphicx} % more modern
\usepackage{subfigure}

\usepackage{natbib}

\usepackage{algorithm}
\usepackage{algorithmic}

\usepackage{hyperref}

\newtheorem{theorem}{Theorem}
\newtheorem{lemma}[theorem]{Lemma}

\newtheorem{remark}[theorem]{Remark}
\newtheorem{definition}[theorem]{Definition}

\usepackage[fleqn]{amsmath}
\usepackage{amssymb}

\usepackage{times}
\usepackage{balance}\usepackage{epsfig}

\usepackage[accepted]{icml2014}

\newcommand\blfootnote[1]{%
  \begingroup
  \renewcommand\thefootnote{}\footnote{#1}%
  \addtocounter{footnote}{-1}%
  \endgroup
}

\icmltitlerunning{An Analysis of Random Projections in Cancelable Biometrics}

\begin{document}

\twocolumn[
\icmltitle{An Analysis of Random Projections in Cancelable Biometrics}

\icmlauthor{Devansh Arpit }{devansha@buffalo.edu}
\icmladdress{SUNY Buffalo, Buffalo, NY, USA}
\icmlauthor{ Ifeoma Nwogu }{inwogu@buffalo.edu}
\icmladdress{SUNY Buffalo, Buffalo, NY, USA}
\icmlauthor{Gaurav Srivastava}{srivastava.g@samsung.com}
\icmladdress{Samsung, USA}
\icmlauthor{Venu Govindaraju}{govind@buffalo.edu}
\icmladdress{SUNY Buffalo, Buffalo, NY, USA}

%\keywords{}
%\maketitle
\vskip 0.3in
]

\begin{abstract}
With increasing concerns about security, the need for highly secure physical biometrics-based authentication systems utilizing \emph{cancelable biometric} technologies is on the rise. Because the problem of cancelable template generation deals with the trade-off between template security and matching performance, many state-of-the-art algorithms successful in generating high quality cancelable biometrics all have random projection as one of their early processing steps. This paper therefore presents a formal analysis of why random projections is an essential step in cancelable biometrics. By formally defining the notion of an \textit{Independent Subspace Structure} for datasets, it can be shown that random projection preserves the subspace structure of data vectors generated from a union of independent linear subspaces. The bound on the minimum number of random vectors required for this to hold is also derived and is shown to depend logarithmically on the number of data samples, not only in independent subspaces but in disjoint subspace settings as well. The theoretical analysis presented is supported in detail with empirical results on real-world face recognition datasets.
\end{abstract}
\vspace{-5mm}
\section{Introduction}\label{sec:intro}
\vspace{-1mm}
As physical biometrics-based authentication such as the use of fingerprints, faces, iris scans etc., has gained significant popularity in last few decades, there is a growing need for \emph{cancelable biometric} technologies. Cancelable biometrics refers to the systematic, intentional, repeatable distortion of biometrics features in order prevent the notion of ``stolen biometrics''. A person's biometrics are stolen for a specific modality, when the feature template used in assigning the biometric to that user is compromised by a masquerading attacker, thus giving the attacker access privileges to the user's resources. Cancelable biometrics are especially important when there is a need to store biometric templates, because if compromised, it is virtually impossible for a user to regenerate the physical traits that were used in creating the templates during enrollment.

\blfootnote{This work was partially funded by the National Science Foundation under the grant number CNS-1314803}

Thus, in an attempt to reduce the vulnerability of such security systems, there has been increased research activity in the areas of cancelable biometrics where the problem deals with the trade-off between template security and matching performance. The state-of-the-art algorithms that have been successful so far in generating high quality cancelable biometrics are all based on random projection \cite{hybridpaper, rmq, GoandNgo}. Of course, the random projection technique alone is not sufficient for generating highly secure and discriminating biometric templates, but is the first fundamental step which occurs before other more complex techniques, such as class-preserving transforms, template hashing, etc., are implemented.

With increasing technological advancements in computational speed and memory, and with increasing volumes of disparate data being collected for security purposes, more high dimensional feature vectors are being used in many biometrics-driven security systems. However, since computational time increases with dimensionality, real-life biometric systems (employing large volumes of high dimensional feature vectors) are
highly susceptible to performance degradation over time. Dimensionality reduction techniques (such as PCA, LDA, LLE \cite{lle}, LPP \cite{lpp}) can be employed to overcome this problem, however, for applications that perform tasks such as generating secure and discriminating biometric templates, where the subspace structure of the data should be preserved after dimensionality reduction, many of these techniques will fail. We foresee the use of random projections as a core component of future security systems using biometric modalities such as face recognition for authentication.

For this reason and more, in this paper, we formally define the notion of a \textit{Independent Subspace Structure} for datasets, and based on this definition, we show that random projection preserves the subspace structure of data vectors generated from a union of independent linear subspaces. Thus the technique can be employed as a cancelable transform to project an original biometric template into a subspace and generate a new cancelable template, while maintaining discriminability. While an extensive number of papers in the literature has employed random projection for data dimensionality reduction for tasks such as k-means clustering \cite{kmeans}, classification \cite{classification2} \cite{classification} etc., these papers have shown that for the respective tasks, certain desired properties of the data vectors are preserved under random projection. However, to the best of our knowledge, a more general and formal analysis of linear subspace structure preservation under random projections has not been reported thus far; this is the main thrust for this paper.
\vspace{-2mm}

\section{Definitions}\label{sec:defn}
A linear subspace in $\mathbb{R}^{n}$ of dimensions $(d)$ can be
represented using a matrix $B\in\mathbb{R}^{n\times d}$ where the columns of $B$ form the support of the subspace. Then any vector in this subspace can be represented as $x=Bw\mspace{10mu}\forall w\in\mathbb{R}^{d}$.
Let there be $K$ independent subspaces denoted by $S_{1},S_{2}, \hdots, S_{K}$.
Any subspace $S_{i}$ is said to be independent of all other subspaces
if there does not exist any non-zero vector in $S_{i}$ which is a
linear combination of vectors in the other subspaces. Formally,
\[
\sum_{i=1}^{K}S_{i}=\oplus_{i=1}^{K}S_{i}
\]
 where, $\oplus$ denotes direct sum of subspaces.

While the above definition states the condition under which two or more subspaces are independent, it does not specifically tells us quantitatively how well they are separated and this leads us to the definition of the margin between a pair of subspaces.

\begin{definition}
\label{def_subspace_margin}
(Subspace Margin) \\
Subspaces $S_{i}$ and $S_{j}$ are separated by margin $\gamma_{ij}$ if
\begin{equation}
\gamma_{ij} = \max_{u \in S_{i},v \in S_{j}} \frac{\langle u,v\rangle}{\lVert u \rVert_{2} \lVert v \rVert_{2}}
\end{equation}
 \end{definition}

Geometrically, the above definition says that margin between
any two subspaces is defined as the maximum dot product between two
unit vectors, one from either subspace. The vector pair $u$ and $v$ that maximize this dot product is known as the \emph{principal vector pair} between the two subspaces while the angle between these vectors is called the \emph{principal angle}. Notice that $\gamma_{ij} \in [0,1]$ such that $\gamma_{ij} =0$ implies that the subspaces are maximally separated while $\gamma_{ij} =1$ implies that the two subspaces are not independent.

Having defined these concepts, our goal is to learn a subspace from any given dataset that is sampled from a union of independent linear subspaces such that this independent subspace structure property is approximately preserved in the dataset. We will make this idea more concrete shortly.

Notice that the above definitions of independent subspaces and separation margin (definition \ref{def_subspace_margin}) apply explicitly to well defined subspaces. So a natural question is: \textit{How do we define these concepts for datasets?} We define the \textit{Independent Subspace Structure} for a dataset as follows,

\begin{definition} (Independent Subspace Structure)\\
Let $X=\{x_{j}\}_{j=1}^{N}$ be a $K$ class dataset of $N$ data vectors in $\mathbb{R}^{n}$ and $X_{i} \subset X$
($i \in\left\{ 1\hdots K\right\}$) such that data vectors in $X_{i} $ belong to
class $i$. Then we say that the dataset $X$ has \textit{Independent Subspace Structure} if each $i^{th}$ class data $x \in X_{i}$ is sampled from a linear subspace $S_{i}$ ($i \in\left\{ 1\hdots K\right\}$) in $\mathbb{R}^{n}$ such that each subspace is independent.
\end{definition}

Again, the above definition only specifies that data samples from different classes belong to independent subspaces. To estimate the margin between subspaces these, we define \textit{Subspace Margin for datasets} as follows:

\begin{definition} (Subspace Margin for datasets) \\
For a dataset $X$ with Independent Subspace Structure, class $i$ ($i \in\left\{ 1\hdots K\right\}$) data is separated from all the other classes with margin $\gamma_{i}$, if $\forall x \in X_{i}$ and $\forall y \in X \setminus \{X_{i} \}$, $\frac{\langle x,y\rangle}{\lVert x \rVert_{2} \lVert y \rVert_{2}} \leq \gamma_{i}$, where $\gamma_{i} \in [0,1)$.
\end{definition}

With these definitions, we will now make the idea of independent subspace structure preservation more concrete.
Specifically, by subspace structure preservation, we refer to the case where we are originally given a set of data vectors sampled from a union of independent linear subspaces and subsequently, after projection, the projected data vectors also belong to a union of independent linear subspaces.

Formally, let $X$ be a $K$ class dataset in $\mathbb{R}^{n}$ with independent subspace structure such that class $i$ samples ($x \in X_{i}$) are drawn from subspace $S_{i}$, then the projected data vectors (using projection matrix $P \in \mathbb{R}^{n \times m}$) in the sets $\bar{X}_{i}:=\{P^{T}x:x\in X_{i}\}$
for $i\in\{1\hdots K\}$ are such that data vectors in each set $\bar{X}_{i}$
belong to a linear subspace ($\bar{S}_{i}$ in $\mathbb{R}^{m}$) and the subspaces $\bar{S}_{i}\mspace{5mu}(i\in\{1\hdots K\})$
are independent, i.e., $\sum_{i=1}^{K}\bar{S}_{i}=\oplus_{i=1}^{K}\bar{S}_{i}$.
\vspace{-3mm}

\section{Random Projections}\label{sec:randproj}
\vspace{-1mm}
Random Projection has gained significant popularity in recent years due to its low computational costs  and the guarantees it comes with. Specifically, it has been shown in cases of linearly separable data \cite{classification} \cite{classification2} and data that lies on a low dimensional compact manifold \cite{manifold1} \cite{manifold2}, that random projection preserves the linear separability and manifold structure respectively, given that certain conditions are satisfied. Notice that a union of independent linear subspaces is a specific case of manifold structure and hence the results of random projection for manifold structure apply in general to our case. However, as those results are derived for a more general case, their results are weak when applied to our problem setting. Further, to the best of our knowledge, there has not been any prior analysis of random projection on the margin between independent subspaces.

The various applications of random projection for dimensionality reduction
are rooted in the following version of the Johnson-Lindenstrauss (JL) lemma
\cite{jl}:

\begin{lemma}
\label{lem_jl}
For any vector x $\in\mathbb{R}^{n}$,
matrix R $\in\mathbb{R}^{m\times n}$ where each element of R is drawn
i.i.d. from a standard Gaussian distribution, $R_{ij}\sim\frac{1}{\sqrt{m}}\mathcal{N}(0,1)$
and any $\epsilon\in(0,1/2)$
\begin{equation}
\begin{split}Pr\left((1-\epsilon)\lVert x\rVert^{2}\leq\lVert Rx\rVert^{2}\leq(1+\epsilon)\lVert x\rVert^{2}\right)\\
\geq1-2\exp\left(-\frac{m}{4}\left(\epsilon^{2}-\epsilon^{3}\right)\right)
\end{split}
\end{equation}
 \end{lemma} This lemma states that the $\ell_{2}$ norm of any
randomly projected vector is approximately equal to the $\ell_{2}$
norm of the original vector. While conventionally, elements
of the random matrix are generated from a Gaussian distribution, it
has been proved \cite{sparseRP} \cite{very_sparse} that one can
indeed use sparse random matrices (with most of the elements being
zero with high probability) to achieve the same goal.

Aside, in relation to adopting random projection in the preliminary steps to providing template cancelability, if given a cancelable biometric template $\bar{x}=Rx$ constructed from an original template $x$ with the projection matrix $R$, and the initial cancelable template $\bar{x}$ is compromised, a new template $\bar{x}' = R'x$ is issued with a new projection matrix $R'$ as a replacement. Lemma \ref{lem_jl} indicates that discriminability of the original feature vector is preserved for each template, however the conditions required for this still need to be investigated.

Before studying the conditions required for independent subspace structure preservation for a multiclass problem, we first state our cosine preservation lemma which simply states that the cosine of angle between any two fixed vectors is approximately preserved under random projection. A similar angle preservation theorem is stated in \cite{classification}, but we will state the difference between the two after presenting the lemma.

\begin{lemma}
\label{lem_cos}
(Cosine preservation) \\
For all $x,y\in\mathbb{R}^{n}$, any $\epsilon\in(0,1/2)$ and
matrix R $\in\mathbb{R}^{m\times n}$ where each element of R is drawn
i.i.d. from a standard Gaussian distribution, $R_{ij}\sim\frac{1}{\sqrt{m}}\mathcal{N}(0,1)$, one of the following inequalities holds true
\begin{equation}
\label{eq_cos1}
\begin{split} \frac{1}{(1-\epsilon)}\frac{\langle x,y\rangle}{\lVert x \rVert_{2} \lVert y \rVert_{2}} - \frac{\epsilon}{1-\epsilon}
 \leq \frac{\langle Rx,Ry\rangle}{\lVert Rx \rVert_{2} \lVert Ry \rVert_{2}} \\
\leq \frac{1}{(1+\epsilon)} \frac{\langle x,y\rangle}{\lVert x \rVert_{2} \lVert y \rVert_{2}}+ \frac{\epsilon}{1+\epsilon}\\
\end{split}
\end{equation}
if $\frac{\langle x,y\rangle}{\lVert x \rVert_{2} \lVert y \rVert_{2}} < -\epsilon$,
\begin{equation}
\label{eq_cos2}
\begin{split} \frac{1}{(1-\epsilon)}\frac{\langle x,y\rangle}{\lVert x \rVert_{2} \lVert y \rVert_{2}} - \frac{\epsilon}{1-\epsilon}
 \leq \frac{\langle Rx,Ry\rangle}{\lVert Rx \rVert_{2} \lVert Ry \rVert_{2}} \\
\leq \frac{1}{(1-\epsilon)} \frac{\langle x,y\rangle}{\lVert x \rVert_{2} \lVert y \rVert_{2}}+ \frac{\epsilon}{1-\epsilon}\\
\end{split}
\end{equation}
if $-\epsilon \leq \frac{\langle x,y\rangle}{\lVert x \rVert_{2} \lVert y \rVert_{2}} < \epsilon$, and
\begin{equation}
\label{eq_cos3}
\begin{split} \frac{1}{(1+\epsilon)}\frac{\langle x,y\rangle}{\lVert x \rVert_{2} \lVert y \rVert_{2}} - \frac{\epsilon}{1+\epsilon}
 \leq \frac{\langle Rx,Ry\rangle}{\lVert Rx \rVert_{2} \lVert Ry \rVert_{2}} \\
\leq \frac{1}{(1-\epsilon)} \frac{\langle x,y\rangle}{\lVert x \rVert_{2} \lVert y \rVert_{2}}+ \frac{\epsilon}{1-\epsilon}\\
\end{split}
\end{equation}
 if $\frac{\langle x,y\rangle}{\lVert x \rVert_{2} \lVert y \rVert_{2}} \geq \epsilon$. Further the inequality holds true with probability at least $1-8\exp\left(-\frac{m}{4}\left(\epsilon^{2}-\epsilon^{3}\right)\right)$.

Proof: See appendix.
\end{lemma}

We would like to point out that cosine of both acute and obtuse angles are preserved under random projection as is evident from the above lemma. However, if the cosine value is close to zero, the additive error in the inequalities \ref{eq_cos1}, \ref{eq_cos2} and \ref{eq_cos3} distorts the cosine significantly after projection. On the other hand, \cite{classification} in their paper state that obtuse angles are not preserved. As evidence, the authors empirically show cosines with negative value close to zero. However, as already stated, cosine values close to zero are not well preserved. Hence this does not serve as an evidence that obtuse angles are not preserved under random projection which we show empirically otherwise to be true. Notice that this is not the case for the JL lemma (\ref{lem_jl}) where the error is multiplicative and hence length of vectors are preserved to a good degree invariantly for all vectors.

In general, the inner product between vectors is not well preserved under random projection irrespective of the angle between the two vectors. This can be analyzed using Equation \ref{eq_inner_notpre}. Rewriting this equation in the following form, we have that,
 \begin{equation}
\label{eq_inner_notpre2}
\begin{split} \langle {x},{y}\rangle -\epsilon \lVert x \rVert_{2} \lVert y \rVert_{2} \leq \langle R{x},R{y}\rangle \leq \langle {x},{y}\rangle +\epsilon \lVert x \rVert_{2} \lVert y \rVert_{2}
\end{split}
\end{equation}
holds with high probability. Clearly, because the error term itself depends on the length of the vectors, inner product between arbitrary vectors after random projection is not well preserved. However, as a special case, inner product of vectors with length less than $1$ is preserved (corollary $2$ in \cite{ref_inner_prod}) because the error term gets diminished in this case.

For ease of representation, in all further analysis, we will use Equation \ref{eq_cos3} while making use of the cosine preservation lemma. We will now go on to examine the conditions under which independent subspace structure can be preserved for any linearly separatable dataset.

\subsection{Subspace Margin Preservation}
In order for independent subspace structure to be preserved for any dataset, we need two conditions to hold simultaneously. First, data sampled from each subspace should continue to belong to a linear subspace after projection. Second, the subspace margin for the dataset should be preserved.

\begin{remark} (Individual Subspace preservation) \\
Let $X_{i}$ denote the set of data vectors ($x$) drawn from the subspace $S_{i}$, and
let $R\in\mathbb{R}^{m\times n}$ denote the random projection matrix
as defined before. Then after projection, all the vectors in $X_{i}$
continue to lie along the linear subspace in the span of $RB_{i}$, where the columns of $B_{i}$ denote the span of $S_{i}$.
\end{remark}

The above straight forward remark states that the first requirement always holds true. Now we need to derive the condition needed for the second requirement to hold true.

\begin{theorem}
\label{th_multi}
 (Multiclass Subspace Preservation) Let $X=\{x_{j}\}_{j=1}^{N}$ be a $K$ class dataset with Independent Subspace structure and the $i^{th}$ class have margin $\gamma_{i}$. Then for any $\epsilon\in(0,1/2)$, the subspace structure of the entire dataset is  preserved after random projection using matrix R $\in\mathbb{R}^{m\times n}$ ($R_{ij}\sim_{i.i.d.} \frac{1}{\sqrt{m}}\mathcal{N}(0,1)$) with margin $\bar{\gamma}_{i}$ for class $i$ as follows
\begin{equation}
\label{eq_th_multi}
\begin{split}
Pr\left(\bar{\gamma}_{i} \leq \frac{1}{(1-\epsilon)} \gamma_{i}+ \frac{\epsilon}{1-\epsilon}, \forall i \in \{1 \hdots K \}\right)\\
\geq1-6N^{2} \exp\left(-\frac{m}{4}\left(\epsilon^{2}-\epsilon^{3}\right)\right)
\end{split}
\end{equation}

Proof: See appendix.
\end{theorem}

%Notice that the probability bounds do not depend on the number of classes in the dataset. This implies that number of classes do not affect subspace structure preservation; it only depends on the number of data vectors.

Recall from our discussions on the cosine preservation lemma (\ref{lem_cos}), that cosine values close to zero are not well preserved under random projection. However, from our above error bound on the margin (eq \ref{eq_th_multi}), it turns out that this is not a problem - two subspaces separated with a margin close to zero implies that the principal angle between them is almost orthogonal, i.e., they are maximally separated. Therefore, under these circumstances, the projected subspaces are also well separated.

Formally, let $\gamma = 0$, so that after projection, $\bar{\gamma} \leq \frac{\epsilon}{1-\epsilon}$ is further upper bounded by $1$ as $\epsilon$ tends to $0.5$. In practice we set $\epsilon$ to be a much smaller quantity, hence $\bar{\gamma}$ is well below $1$.

While the analysis so far only relates to structure preservation for datasets with independent subspace structure, it is not hard to see that the same bounds also apply to datasets with disjoint subspace structure, i.e., each subspace (class) is pairwise disjoint with each other but not independent overall.
\vspace{-1mm}

\section{Sparse Representation based Recognition}\label{sec:sparserep}
Sparse representation (SR) has been widely used for classification purposes in various machine learning applications, including face recognition tasks in biometric security applications. The idea of SR is based on the theory of compressed sensing. This theory claims that if a system of linear equations with an overcomplete dictionary has a sparse solution then it can be achieved by solving the basis pursuit algorithm:

\begin{equation}
\label{bas_pur}
w^{*} = \arg \min_{w}  \lVert w \rVert_{1} \mspace{10mu} s.t. \mspace{10mu} y=Dw
\end{equation}
where $y \in \mathbb{R}^{m}$ is the measurement vector, $D \in \mathbb{R}^{m \times n}$ is overcomplete dictionary and $w \in \mathbb{R}^{n}$ is the variable for which we want a sparse solution. This property is very useful for classification because one can use all the training samples as the columns of the overcomplete dictionary $D$, test sample as $y$ and solve the above optimization to obtain the sparse reconstruction coefficient $w^{*}$ over the training samples. The advantage of representing a test sample as a sparse linear combination of the training samples is that fewer non-zero coefficients over the training samples will be more discriminative in terms of the class of the test sample.

More recently, Sparse Subspace clustering (SSC) has been used for subspace clustering applications. The subspace clustering domain assumes that each individual class lies along a linear independent subspace and under this assumption we want to cluster a given set of data samples such that each cluster corresponds to samples from one such subspace. The authors of SSC approach \cite{SSC} show that, the basis pursuit optimization guarantees the correct reconstruction of a test sample ($y$) using an overcomplete dictionary of training samples ($D$). Formally this is stated in the following theorem:

\begin{theorem} (Theorem 1 in \cite{SSC})\\
Let $D \in \mathbb{R}^{m \times n}$ be a matrix whose columns are drawn from a union of $K$ independent linear subspaces. Assume that the points within each subspace are in general position. Let $y$ be a new point in subspace $i$. The solution to the $\ell^{1}$ problem in \ref{bas_pur}, $w^{*} \in \mathbb{R}^{n}$ is sparse such that $w_{j}\neq 0$ iff $D_{j}$ belongs to the $i^{th}$ subspace and $w_{j}$=0 otherwise.
\end{theorem}

where $D_{j}$ denotes the $j^{th}$ column of matrix $D$.
This theorem gives us the sufficient condition under which one is guaranteed to recover the correct coefficients for a given test sample using SR. This property is used in the SSC algorithm for clustering. However, this also clearly shows why it makes sense to use sparse representation for the task of classification under the assumption that our classes lie along independent linear subspaces. This assumption is widely used for applications like face recognition and motion segmentation.

\begin{figure}[t]
\centering{ \subfigure[$\epsilon=0.1$]{\includegraphics[width=0.3\columnwidth]{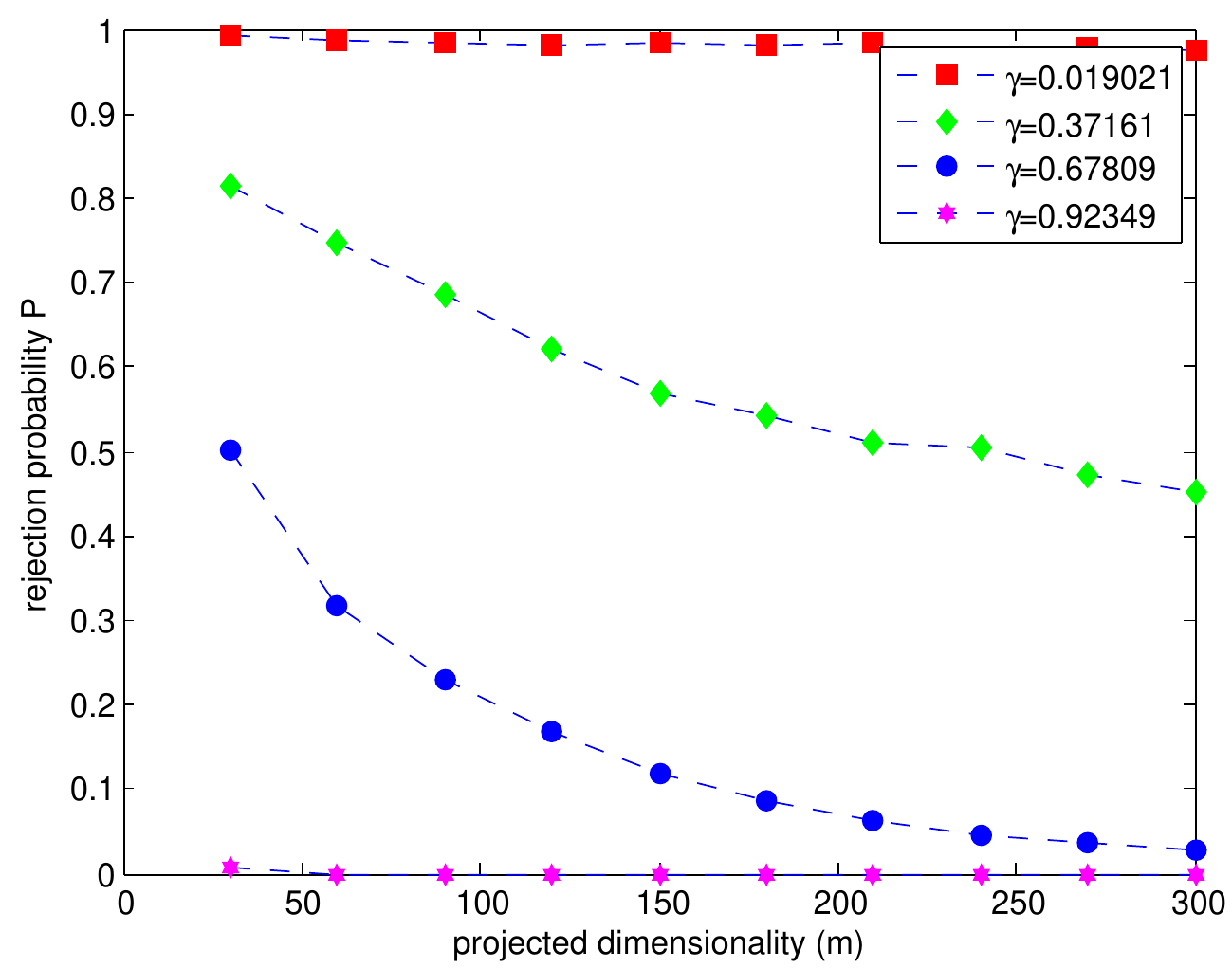}}
\hspace{0.1in} \subfigure[$\epsilon=0.3$]{\includegraphics[width=0.3\columnwidth]{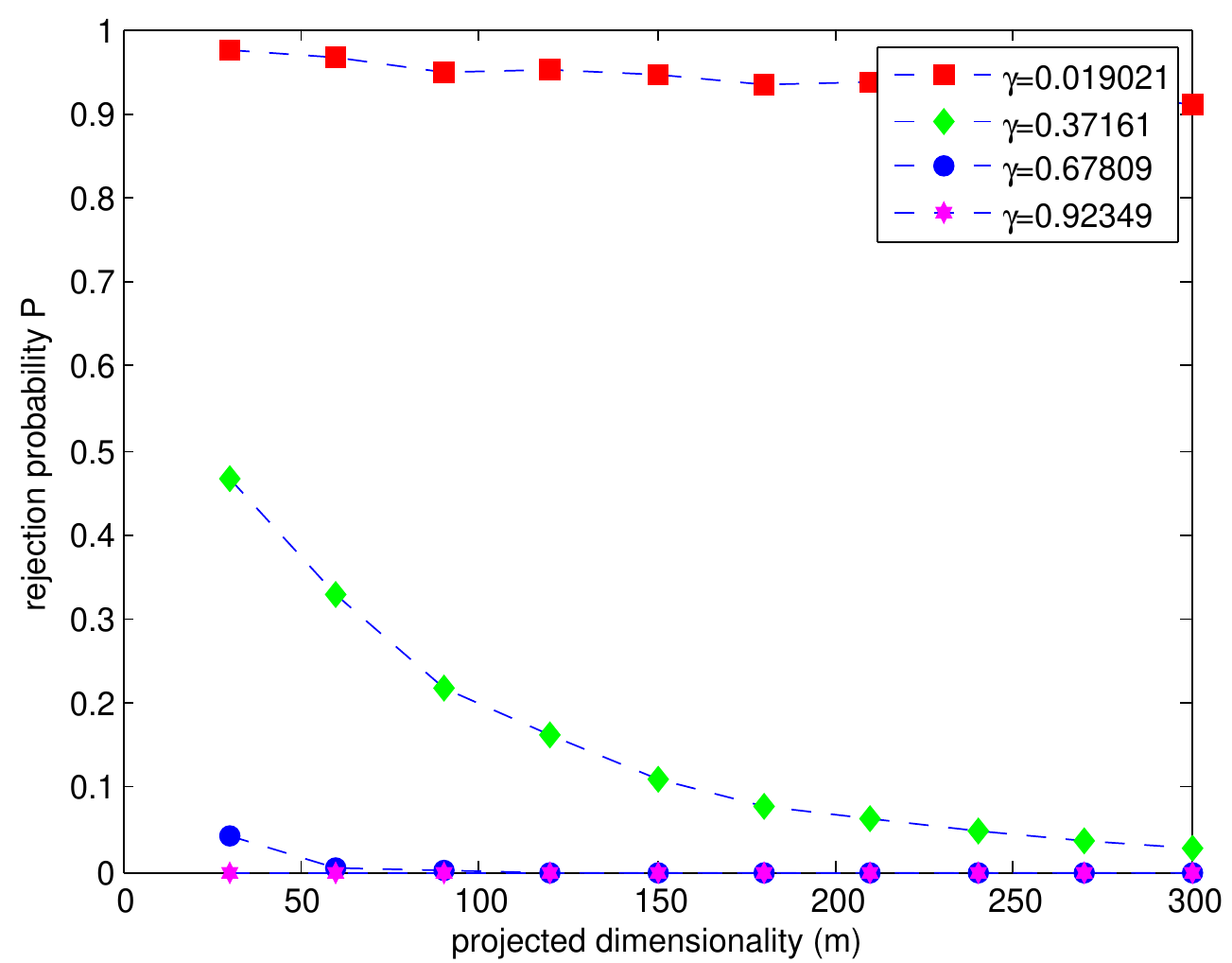}}\\
\subfigure[$\epsilon=0.1$]{\includegraphics[width=0.3\columnwidth]{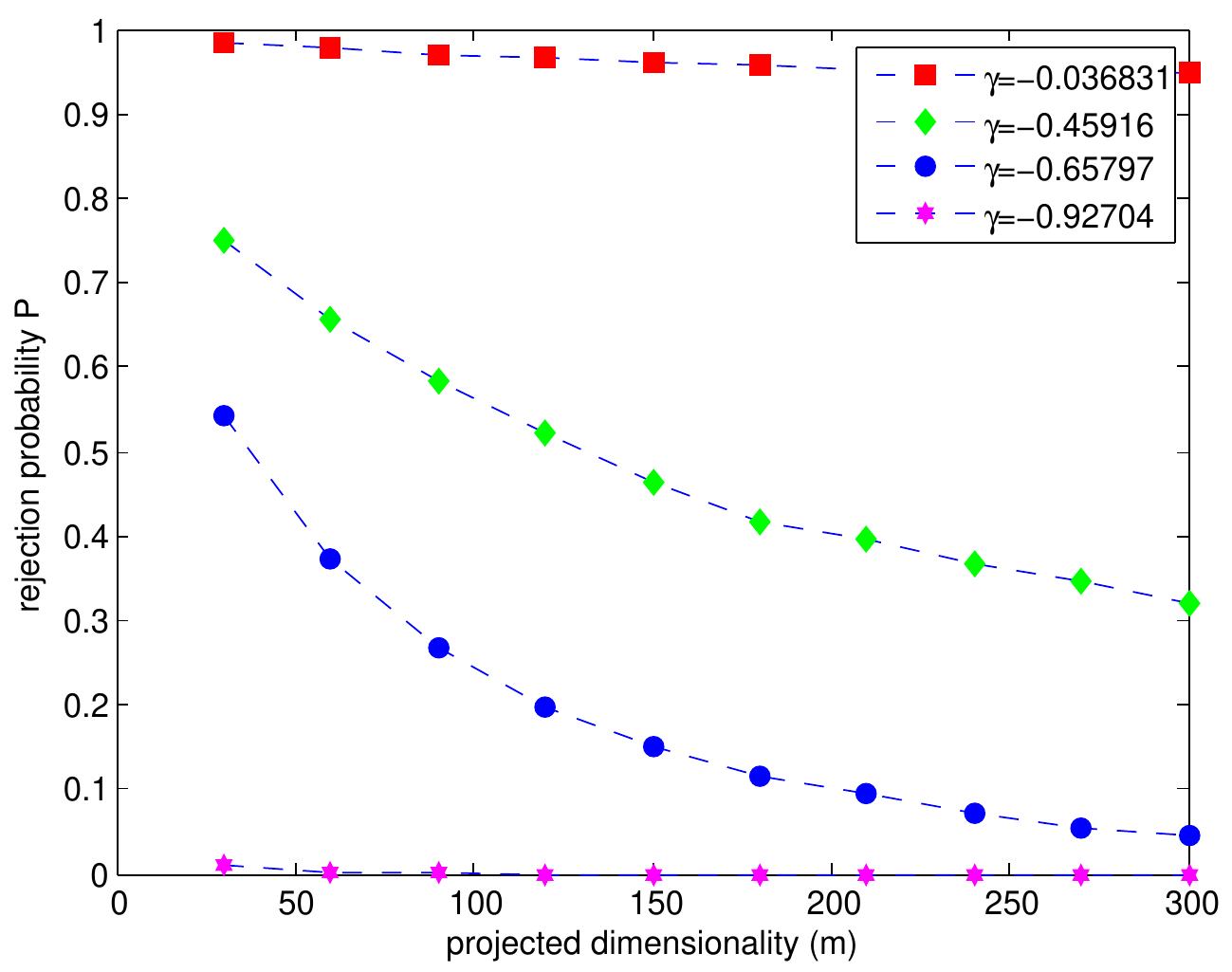}}
\hspace{0.1in} \subfigure[$\epsilon=0.3$]{\includegraphics[width=0.3\columnwidth]{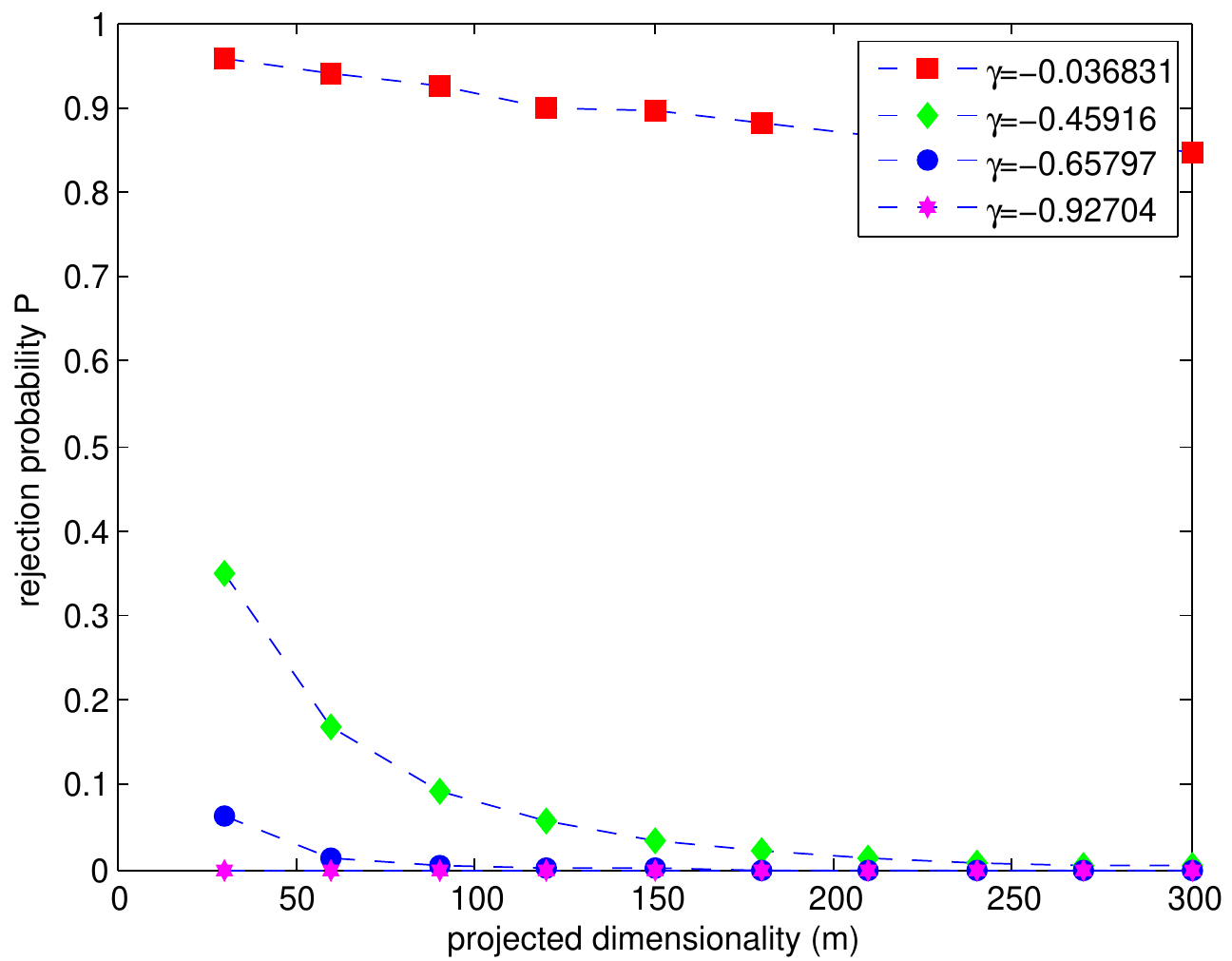}}
 \caption{\textbf{Empirical rejection probability for cosine preservation of acute (a,b) and obtuse angles (c,d).} See section \ref{sec_reject_angle} for details. \label{fig_reject_angle}
}}
\end{figure}

\begin{figure}[t]
\centering{ \subfigure[$\epsilon=0.1$]{\includegraphics[width=0.3\columnwidth]{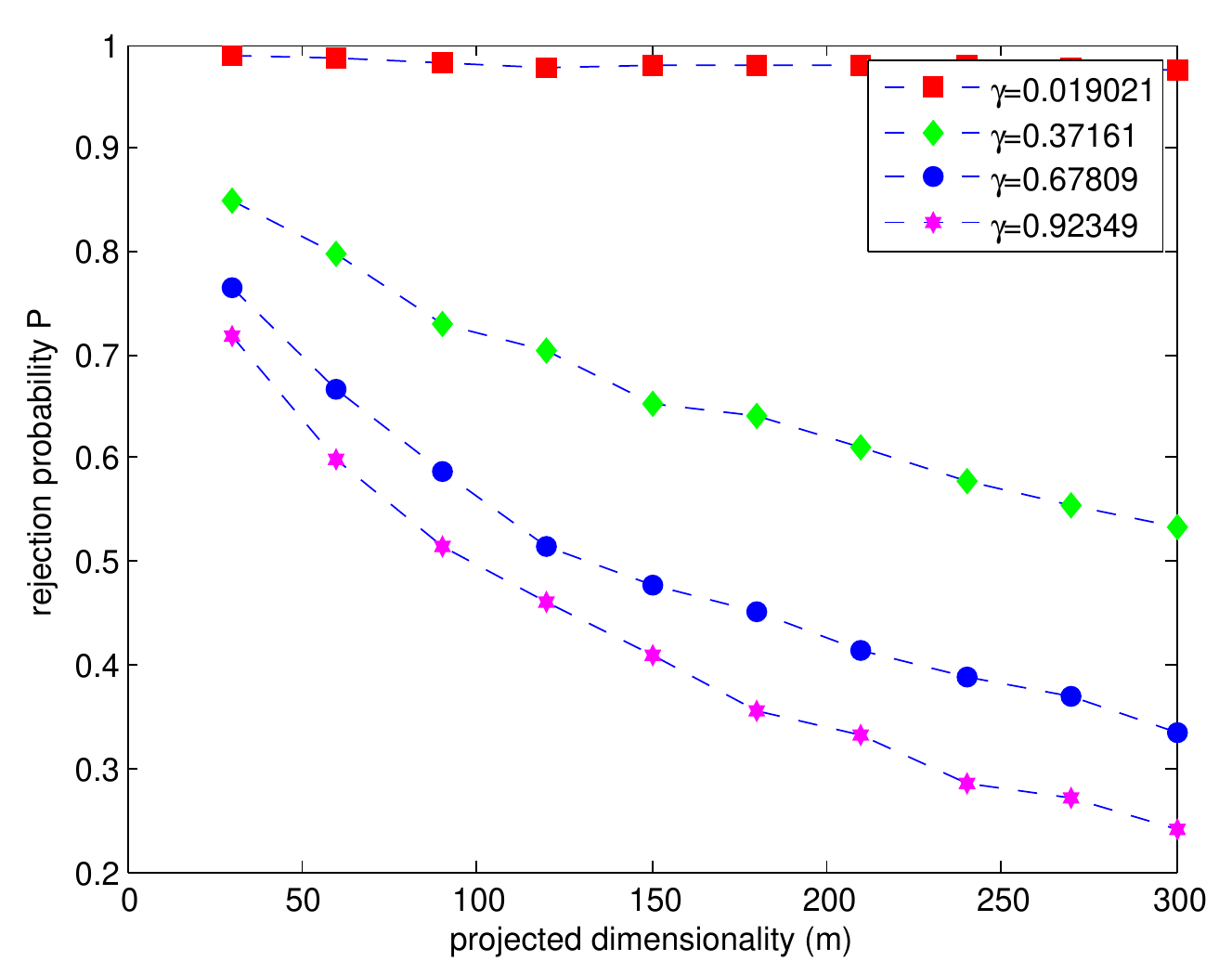}}
\hspace{0.1in} \subfigure[$\epsilon=0.3$]{\includegraphics[width=0.3\columnwidth]{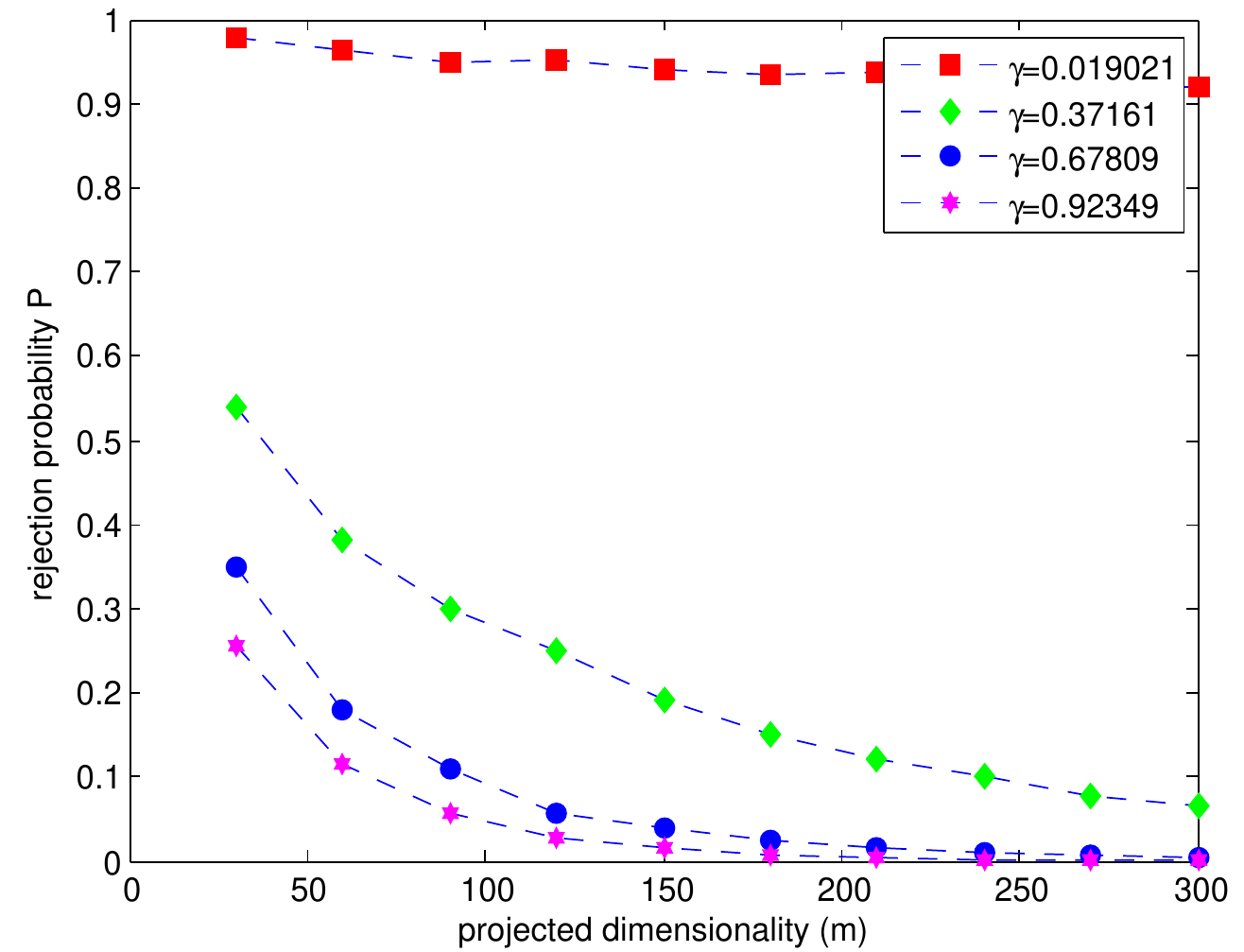}}\\
\subfigure[$\epsilon=0.1$]{\includegraphics[width=0.3\columnwidth]{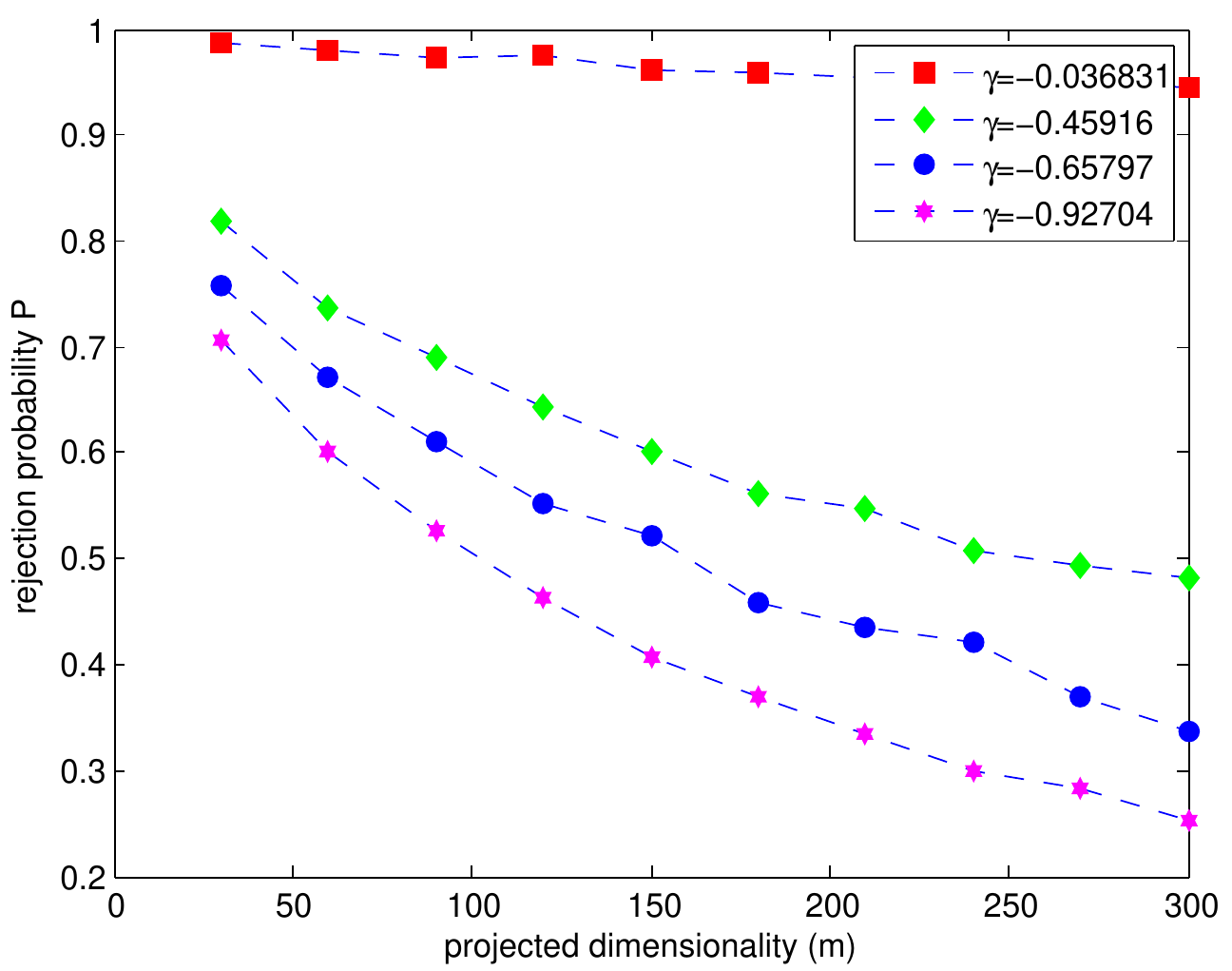}}
\hspace{0.1in} \subfigure[$\epsilon=0.3$]{\includegraphics[width=0.3\columnwidth]{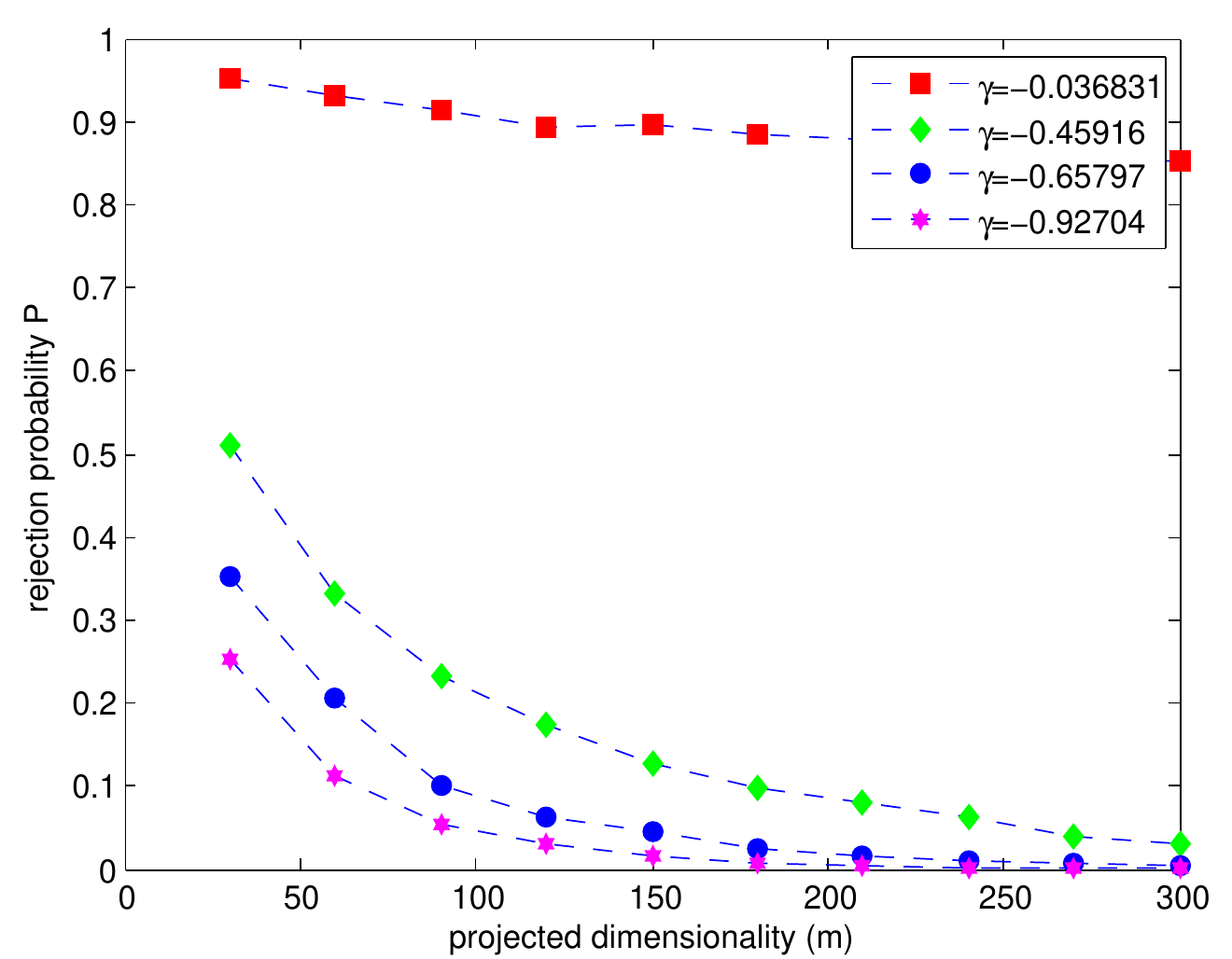}}
 \caption{\textbf{Empirical rejection probability for inner product preservation of acute (a,b) and obtuse angles (c,d).} See section \ref{sec_reject_inner} for details. \label{fig_reject_inner}
}}
\vspace{-5mm}
\end{figure}

Since the above algorithms make use of the underlying subspace assumption for datasets, it is natural to investigate if there exists a dimensionality reduction method that is guaranteed to preserve this structure in the dataset. If so, we can apply the aforementioned algorithms in a much smaller feature space without losing accuracy while simultaneously being much faster.

In the preceding section, we showed that random projection preserves the underlying structure in datasets and thus can be effectively used for dimensionality reduction. Notice that the advantage of random projections is three fold: it allows for the classification/recognition algorithm to run faster; (ii) it is  extremely inexpensive to compute; and (iii) it yields classification results with accuraies at par with that in the original dimensions of the data. While most dimensionality reduction algorithms are expensive in terms of computing the projection vectors (e.g. PCA takes cubic time in the size of feature space), random projection needs each element of its projection vectors to be sampled randomly independent of the data at hand. This non-adaptive nature of random projection makes it a very powerful dimensionality reduction tool. These qualities indicate why random projections is becoming such an essential technique for developing very efficient and highly secure biometric applications.

\section{Empirical Analysis}
\vspace{-1mm}
In this section, we present empirical evidence to support our theoretical analysis (from Section \ref{sec:randproj}) of why random projections work for cancelable biometrics. We perform experiments to show both cosine preservation and subspace structure preservation under random projections, using different face recognition datasets.

\subsection{Cosine and Inner product Preservation}
\vspace{-1mm}
\subsubsection{Cosine preservation}
\label{sec_reject_angle}
\vspace{-1mm}
In lemma \ref{lem_cos}, we concluded that the cosine of the angle between any two vectors remains preserved under random projection irrespective of the angle being acute or obtuse. However, we also stated that cosine values close to zero are not well preserved. Here, we perform empirical analysis on vectors with varying angles (both acute and obtuse) and arbitrary length to verify the same. In order to achieve this, we use settings similar to \cite{classification}. We generate $2000$ random projection matrices $R_{i} \in \mathbb{R}^{m \times n}$ ($i=1$ to $2000$) where we vary $m = \{ 30,60, \hdots 300 \}$ and $n=300$ is the dimension of the original space. We define empirical rejection probability for cosine preservation similar to \cite{classification} as,
\begin{equation}
\label{eq_reject_angle}
\begin{split}
\hat{P} = 1-\frac{1}{2000}\sum_{i=1}^{2000} \textbf{1} ( (1-\epsilon) \leq \frac{\langle R_{i}x,R_{i}y\rangle \lVert x \rVert_{2} \lVert y \rVert_{2}}{\lVert R_{i}x \rVert_{2} \lVert R_{i}y \rVert_{2} \langle x,y\rangle} \\
\leq (1+\epsilon ) ) \nonumber
\end{split}
\end{equation}
where we vary $\epsilon \in \{ 0.1,0.3 \}$ and $\textbf{1}(.)$ is the indicator operator.

\begin{table}
\caption{\textbf{Time taken (in msec) for dimensionality reduction on Extended Yale dataset B.} See section \ref{sec_subspace_preservation} for details. \label{table:dim_red_yaleb}}
\centering{}%
\begin{tabular}{|c|c|c|c|c|}
\hline
method/dimensions & 100  & 150  & 300  & 500\tabularnewline
\hline
RP  & 1.59  & 2.76  & 11.11  & 25.78\tabularnewline
\hline
PCA  & 322.04 & 328.58  & 332.64  & 341.10 \tabularnewline
\hline
\end{tabular}
\vspace{-5mm}
\end{table}

\begin{table}
\caption{\textbf{Classification accuracy on Extended Yale dataset B.} See section \ref{sec_subspace_preservation} for details. \label{table:yale_classification}}
\centering{}%
\begin{tabular}{|c|c|c|c|c|}
\hline
method/dimensions  & 100  & 150  & 300  & 500\tabularnewline
\hline
RP  & \textbf{94.61}  & \textbf{95.85}  & \textbf{96.53}  & \textbf{97.93}\tabularnewline
\hline
PCA  & 94.27 & 94.95  & 95.94  & 96.86 \tabularnewline
\hline
\end{tabular}
\vspace{-5mm}
\end{table}

For acute angle, we randomly generate vectors $x$ and $y$ of arbitrary length but with fixed cosine values $\gamma=\{0.019021, 0.37161, 0.67809, 0.92349 \}$. For obtuse angle, we similarly generate vectors $x$ and $y$ with fixed cosine values $\gamma=\{-0.036831, -0.45916, -0.65797, -0.92704\}$. We then compute the empirical rejection probability as mentioned above for different values of $\epsilon$. Figure \ref{fig_reject_angle} shows the results on these vectors. In the figure, notice that the rejection probability decreases as the absolute value of cosine of the angle ($\gamma$) increases (from $0$ to $1$), as well as for higher value of $\epsilon$. Notice, for cosine values close to zero, the rejection probability is close to $1$ even at high dimensions. These results corroborate with our theoretical analysis in lemma \ref{lem_cos}.

\subsubsection{Inner Product under Random projection}
\label{sec_reject_inner}
\vspace{-1mm}
We use the same experimental setting as in section \ref{sec_reject_angle}. We define the empirical rejection probability of inner product similar to \cite{classification} as
\begin{equation}
\label{eq_reject_inner}
\begin{split}
\hat{P} = 1-\frac{1}{2000}\sum_{i=1}^{2000} \textbf{1} ( (1-\epsilon) \leq \frac{\langle R_{i}x,R_{i}y\rangle}{ \langle x,y\rangle}
\leq (1+\epsilon ) ) \nonumber
\end{split}
\end{equation}

We use the same vectors as in \ref{sec_reject_angle} for experiments in this section. We then compute the empirical rejection probability as mentioned above for different values of $\epsilon$. Figure \ref{fig_reject_inner} shows the results on these vectors. As is evident from the figure, inner product between vectors is not well preserved (even when cosine values are close to $1$). This result is in line with our theoretical bound in equation \ref{eq_inner_notpre2} as the vector lengths in our experiment are arbitrarily greater than $1$.

\begin{table}
\caption{\textbf{Time taken (in msec) for dimensionality reduction on PIE dataset} See section \ref{sec_subspace_preservation} for details. \label{table:dim_red_pie}}
\centering{}%
\begin{tabular}{|c|c|c|c|c|}
\hline
method/dimensions & 30  & 50  & 70  & 100\tabularnewline
\hline
RP  & 0.35  & 0.60  & 0.73  & 1.3\tabularnewline
\hline
PCA  & 317.3 & 315.9  & 318.8  & 319.6 \tabularnewline
\hline
\end{tabular}
\vspace{-5mm}
\end{table}

\begin{table}
\caption{\textbf{Classification accuracy on PIE dataset.} See section \ref{sec_subspace_preservation} for details. \label{table:pie_classification}}
\centering{}%
\begin{tabular}{|c|c|c|c|c|}
\hline
method/dimensions  & 30  & 50  & 70  & 100\tabularnewline
\hline
RP  & {96.67}  & \textbf{97.45}  & \textbf{98.04}  & \textbf{98.04}\tabularnewline
\hline
PCA  & \textbf{97.06} & 97.45  & 97.45  & 97.25 \tabularnewline
\hline
\end{tabular}
\vspace{-5mm}
\end{table}

\subsubsection{Required number of random vectors}
\label{sec_req_rp}
\vspace{-1mm}
We study the number
of random vectors required for subspace preservation by varying different
parameters. The lower bound on the number of random vectors required for theorem \ref{th_multi} to hold is given by,
\begin{equation}
m\geq\frac{8}{(\epsilon^{2}-\epsilon^{3})}\mspace{5mu}\ln \frac{\sqrt{6}N}{1-\delta}
\end{equation}
It can be seen that
for $N=1000$ and $\epsilon=0.15$, random projection to lower dimensions is effective
only if $m>6000$ while for $\epsilon=0.4$, $m>900$ suffices. The
choice of $\epsilon$ depends on the robustness of the algorithm (for
the respective task) towards noise and is a trade-off between noise
(allowed) and the number of random vectors ($m$) required.

\subsection{Subspace Structure preservation}
\label{sec_subspace_preservation}
\vspace{-1mm}
In this section, our goal is to show that random projections achieve accuracy better or at least at par with the most widely used dimensionality reduction technique (PCA). We report comparative analysis on the accuracy results and performance times between random projections and PCA. We selected PCA alone for detailed analysis mainly because we found the performance of the other nonlinear dimensionality reduction techniques to be significantly less than the two techniques. Testing on the Extended Yale dataset B (described below), we initially used LPP (Locality Preserving Projections), NPE (Neighborhood Preserving Embedding) \cite{npe}, and Laplacian Eigenmaps \cite{Belkin} to reduce the data to 150 dimensions. The best performing of these reduction techniques yielded a result of only 73\% compared to the close to 96\% accuracies resulting from random projections and PCA. In fairness, these other techniques make no claim to preserving the original subspace structure of the data, rather they preserve some general manifold structure, and do not necessarily guarantee subspace separability.

With this intent of showing that random projections achieve accuracy better or at least at par with PCA, we use sparse representation based classification (SRC, \cite{src}) technique that exploits the subspace structure in the data. One can always use a better classification algorithm that exploits this structure to achieve higher accuracy. However, our aim here is not to compare different classification algorithms but to show that random projection is a computationally inexpensive dimensionality reduction tool with performance guarantees supported by our theoretical analysis.

Cancelable biometrics on face templates is our testbed-of-choice because it is generally assumed that face images with illumination variation lie along linear
independent subspaces \cite{face_subspace}. We use the following datasets for evaluation:

\textit{1. Extended Yale dataset B \cite{yaleb}}: It consists of $\sim2414$ frontal face images of 38
individuals ($K=38$) with $64$ images per person. These images were taken under constrained but varying illumination conditions. We crop all the images to $32 \times 32$ and concatenate all the pixel intensity to form our feature vectors. We use a $50 \% - 50 \%$ train-test split for evaluation.

\textit{2. PIE dataset \cite{pie}}: The CMU pose, illumination, and expression (PIE) database consists of $41368$ images of $68$ people ($K=68$) under $13$ different poses, $43$ illumination conditions and $4$ different expressions. However, we utilize only the first 10 classes from this dataset with $70 \% - 30 \%$ train-test split for evaluation. We cropped to size $32 \times 32$ pixels. The pixel intensities are concatenated to form the feature vectors.

We perform two types of experiments. First, we compare the time taken for dimensionality reduction by PCA and random projections for both datasets. This time is the sum of the time taken by either algorithm to compute it's projection vectors and then to project the entire dataset down to these projection vectors. The results are shown in Table \ref{table:dim_red_yaleb} and \ref{table:dim_red_pie} for the Extended Yale B dataset and PIE dataset respectively. The results show that random projections is faster than PCA by at least an order of $10$ times.

Secondly, we show classification accuracies on both the datasets after dimensionality reduction. These results are shown in Table \ref{table:yale_classification} and \ref{table:pie_classification} for Extended Yale B dataset and PIE dataset respectively. Clearly, random projections performs better than PCA while being significantly faster.

These results substantiate our claim that random projections preserve the subspace structure of any given dataset. Also notice that even a very low number of random vectors used for projection yields good accuracy. This observation can be explained using Lemma $10$ of \cite{dlogd} where the authors show that if a given data lies along a $d$ dimensional subspace then one only needs $O(d \log d)$ random vectors. In most real applications the value of $d$ is usually low, i.e., classes usually lie along a low dimensional subspace. Thus it is not surprising that even small number of random vectors yield high accuracy.

\section{Discussion}
A major advantage of random projections occurs for streaming data where $N$ is constantly changing. Also, as long as the data lies in a $d$-dimensional subspace, as stated in Lemma $10$ of \cite{dlogd}, $O(d \log d)$ random projection vectors preserve the length of all the vectors in that subspace, hence our structure preservation results still hold true. Thus our results not only hold true for a fixed size dataset, but also for an infinite stream of data vectors, as long as a sufficient (but finite and small, $O(d \log d)$) number of random vectors are used and the underlying data structure remains the same.

As originally stated in Section \ref{sec:intro}, the random projections technique by itself is not a complete solution to generating highly secure and discriminating biometric templates. Although the random projection step is secure against the brute-force attack because original templates are often real-valued and high-dimensional, if the projection matrix is not well protected, an attacker could construct its pseudo-inverse to recover an approximation
to the original data. Nevertheless, with the advantages of random projections namely: allowing for the classification/recognition algorithm to run faster; (ii) being extremely inexpensive to compute; and (iii) yielding classification results with accuracies at par with that in the original dimensions of the data, random projections is quickly becoming an essential early-step technique in the development of very efficient and highly secure biometric applications.

\section{Conclusion}
\vspace{-1mm}
In this paper, we presented a formal analysis of why random projections are an essential initial step for generating cancelable biometrics, especially in a real-life scenario where security, discriminability and cancelability are required. Using random projections for dimensionality reduction ensures that the independent subspace structure of datasets are preserved. We derived the bound on the minimum number of random vectors required for this to hold (Section \ref{sec_req_rp}) and concluded that this number depends logarithmically on the number of data samples. All the above arguments hold under \textit{disjoint subspace} settings as well. As a side analysis, we also showed that while cosine values (lemma \ref{lem_cos})are preserved under random projection for both acute and obtuse angles, inner product (equation \ref{eq_inner_notpre2}) between vectors are not well preserved in general.

Although we describe our work in the context of cancelable biometrics, the discussion and evaluations presented is a detailed analysis of linear subspace structure preservation under random projections, irrespective of the task-at-hand.

\bibliographystyle{icml2014}
\bibliography{Subspace_learning}
\newpage
\input{Subspace_learning_appendix.tex}
\end{document}

%% file: Subspace_learning_appendix.tex
\appendix    
\appendixpage
%\addappheadtotoc\textbf{}
\begin{appendix}
%\bigg{Appendix} 
\section{Proof of Lemma 5}\label{app_th7}
Let $\bar{x}:=x/\lVert x \lVert_{2}$ and $\bar{y}:=y/\lVert y \lVert_{2}$ and consider the case when $\frac{\langle x,y\rangle}{\lVert x \rVert_{2} \lVert y \rVert_{2}} \geq \epsilon$. Then from lemma \ref{lem_jl},
\begin{equation}
\label{eq_pm}
\begin{split}Pr\left((1-\epsilon)\lVert \bar{x}+\bar{y}\rVert^{2}\leq\lVert R\bar{x}+R\bar{y}\rVert^{2}\leq(1+\epsilon)\lVert \bar{x}+\bar{y}\rVert^{2}\right)\\
\geq1-2\exp\left(-\frac{m}{4}\left(\epsilon^{2}-\epsilon^{3}\right)\right)\\
Pr\left((1-\epsilon)\lVert \bar{x}-\bar{y}\rVert^{2}\leq\lVert R\bar{x}-R\bar{y}\rVert^{2}\leq(1+\epsilon)\lVert \bar{x}-\bar{y}\rVert^{2}\right)\\
\geq1-2\exp\left(-\frac{m}{4}\left(\epsilon^{2}-\epsilon^{3}\right)\right)
\end{split}
\end{equation}

Using union bound on the above two, both hold true simultaneously
with probability $1-4\exp\left(-\frac{m}{4}\left(\epsilon^{2}-\epsilon^{3}\right)\right)$.
Notice that $\lVert R\bar{x}+R\bar{y}\rVert^{2}-\lVert R\bar{x}-R\bar{y}\rVert^{2}=4\langle R\bar{x},R\bar{y}\rangle$.
Using \ref{eq_pm}, we get
\begin{equation}
\label{eq_unorm_cos}
\begin{split}4\langle R\bar{x},R\bar{y}\rangle=\lVert R\bar{x}+R\bar{y}\rVert^{2}-\lVert R\bar{x}-R\bar{y}\rVert^{2}\\
\leq(1+\epsilon)\lVert \bar{x}+\bar{y}\rVert^{2}-(1-\epsilon)\lVert \bar{x}-\bar{y}\rVert^{2}\\
=(1+\epsilon)(2+2\langle \bar{x},\bar{y}\rangle)-(1-\epsilon)(2-2\langle \bar{x},\bar{y}\rangle)\\
=4\epsilon+4\langle \bar{x},\bar{y}\rangle
\end{split}
\end{equation}
We can similarly prove in the other direction to yield $\langle R\bar{x},R\bar{y}\rangle\geq\langle \bar{x},\bar{y}\rangle-\epsilon$. Together we have that
 \begin{equation}
\label{eq_inner_notpre}
\begin{split} \langle \bar{x},\bar{y}\rangle -\epsilon \leq \langle R\bar{x},R\bar{y}\rangle \leq \langle \bar{x},\bar{y}\rangle +\epsilon
\end{split}
\end{equation}
holds true with probability at least $1-4\exp\left(-\frac{m}{4}\left(\epsilon^{2}-\epsilon^{3}\right)\right)$.

Finally, applying lemma \ref{lem_jl} on vectors $x$ and $y$, we get
\begin{equation}
\begin{split}
(1 - \epsilon)\lVert x \rVert_{2} \lVert y \rVert_{2} \leq \lVert Rx \rVert_{2} \lVert Ry \rVert_{2}
\end{split}
\end{equation}
Thus, $\langle R\bar{x},R\bar{y}\rangle = \frac{\langle Rx,Ry \rangle}{\lVert x \rVert_{2} \lVert y \rVert_{2}} \geq (1-\epsilon) \frac{\langle Rx,Ry \rangle}{\lVert Rx \rVert_{2} \lVert Ry \rVert_{2}}$. Combining this with eq \ref{eq_inner_notpre}, we get
 $ \frac{\langle Rx,Ry \rangle}{\lVert Rx \rVert_{2} \lVert Ry \rVert_{2}} \leq \frac{1}{(1-\epsilon)} \langle \bar{x},\bar{y}\rangle+ \frac{\epsilon}{1-\epsilon}$. We can similarly get the other inequality to achieve \ref{eq_cos1}. Notice that we made use of lemma \ref{lem_jl} four times and hence inequality \ref{eq_cos1} holds with probability at least $1-8\exp\left(-\frac{m}{4}\left(\epsilon^{2}-\epsilon^{3}\right)\right)$ using union bound.

Inequalities \ref{eq_cos2} and \ref{eq_cos3} can be achieved similarly. $\square$

\section{Proof of Theorem 7}\label{app_th7}
Applying union bound on lemma \ref{lem_cos} for a single vector $x \in X_{1}$ and all vectors $y \in X \setminus \{X_{1}\}$,
\begin{equation}
\label{eq_multi1}
\begin{split}  \frac{\langle Rx,Ry\rangle}{\lVert Rx \rVert_{2} \lVert Ry \rVert_{2}}
\leq \frac{1}{(1-\epsilon)} \gamma_{i}+ \frac{\epsilon}{1-\epsilon}
\end{split}
\end{equation}
holds with probability at least $(1-6\bar{N}_{2} \exp\left(-\frac{m}{4}\left(\epsilon^{2}-\epsilon^{3}\right)\right))$, where $\bar{N}_{i} := N-N_{i}$. Again, applying the above bound for all the samples $x \in X_{1}$,
\begin{equation}
\label{eq_multi2}
\begin{split}  \frac{\langle Rx,Ry\rangle}{\lVert Rx \rVert_{2} \lVert Ry \rVert_{2}}
\leq \frac{1}{(1-\epsilon)} \gamma_{i}+ \frac{\epsilon}{1-\epsilon}
\end{split}
\end{equation}
holds with probability at least $(1-6N_{1}\bar{N}_{1} \exp\left(-\frac{m}{4}\left(\epsilon^{2}-\epsilon^{3}\right)\right))$. Computing bounds similar to \ref{eq_multi2} for all the classes, we have that,
\begin{equation}
\label{eq_multi3}
\begin{split}  \bar{\gamma}_{i}
\leq \frac{1}{(1-\epsilon)} \gamma_{i}+ \frac{\epsilon}{1-\epsilon},  \forall i \in \{1 \hdots K \}
\end{split}
\end{equation}
holds with probability at least $(1- \sum_{i=1}^{K} 6N_{i}\bar{N}_{i} \exp\left(-\frac{m}{4}\left(\epsilon^{2}-\epsilon^{3}\right)\right))$. Notice that $\sum_{i=1}^{K} N_{i}\bar{N}_{i} \leq \sum_{i=1}^{K} N_{i}N = N^{2}$ which leads to \ref{eq_th_multi}. $\square$
\end{appendix}

%% file: Subspace_learning.bbl
\begin{thebibliography}{22}
\providecommand{\natexlab}[1]{#1}
\providecommand{\url}[1]{\texttt{#1}}
\expandafter\ifx\csname urlstyle\endcsname\relax
  \providecommand{\doi}[1]{doi: #1}\else
  \providecommand{\doi}{doi: \begingroup \urlstyle{rm}\Url}\fi

\bibitem[Achlioptas(2003)]{sparseRP}
Achlioptas, Dimitris.
\newblock Database-friendly random projections: Johnson-lindenstrauss with
  binary coins.
\newblock \emph{J. Comput. Syst. Sci.}, 66\penalty0 (4):\penalty0 671--687,
  June 2003.

\bibitem[Arriaga \& Vempala(2006)Arriaga and Vempala]{ref_inner_prod}
Arriaga, Rosa~I. and Vempala, Santosh.
\newblock An algorithmic theory of learning: Robust concepts and random
  projection.
\newblock \emph{Machine Learning}, 63:\penalty0 161--182, 2006.

\bibitem[Balcan et~al.(2004)Balcan, Blum, and Vempala]{classification2}
Balcan, Maria-Florina, Blum, Avrim, and Vempala, Santosh.
\newblock Kernels as features: On kernels, margins, and low-dimensional
  mappings.
\newblock In \emph{In 15th International Conference on Algorithmic Learning
  Theory (ALT ’04}, pp.\  79--94, 2004.

\bibitem[Baraniuk \& Wakin(2009)Baraniuk and Wakin]{manifold1}
Baraniuk, Richard~G. and Wakin, Michael~B.
\newblock Random projections of smooth manifolds.
\newblock \emph{Foundations of Computational Mathematics}, 9:\penalty0 51--77,
  2009.

\bibitem[Belkin \& Niyogi(2003)Belkin and Niyogi]{Belkin}
Belkin, Mikhail and Niyogi, Partha.
\newblock Laplacian eigenmaps for dimensionality reduction and data
  representation.
\newblock \emph{Neural Comput.}, 15\penalty0 (6):\penalty0 1373--1396, June
  2003.
\newblock ISSN 0899-7667.

\bibitem[Boutsidis et~al.(2010)Boutsidis, Zouzias, and Drineas]{kmeans}
Boutsidis, Christos, Zouzias, Anastasios, and Drineas, Petros.
\newblock Random projections for $k$-means clustering.
\newblock \emph{CoRR}, abs/1011.4632, 2010.

\bibitem[Elhamifar \& Vidal(2009)Elhamifar and Vidal]{SSC}
Elhamifar, Ehsan and Vidal, Ren{\'e}.
\newblock Sparse subspace clustering.
\newblock In \emph{CVPR}, pp.\  2790--2797, 2009.

\bibitem[Feng et~al.(2010)Feng, Yuen, and Jain]{hybridpaper}
Feng, Yi~Cheng, Yuen, Pong~Chi, and Jain, Anil~K.
\newblock A hybrid approach for generating secure and discriminating face
  template.
\newblock \emph{IEEE Transactions on Information Forensics and Security},
  5\penalty0 (1):\penalty0 103--117, 2010.

\bibitem[Georghiades et~al.(2001)Georghiades, Belhumeur, and Kriegman]{yaleb}
Georghiades, A.S., Belhumeur, P.N., and Kriegman, D.J.
\newblock From few to many: Illumination cone models for face recognition under
  variable lighting and pose.
\newblock \emph{IEEE Trans. Pattern Anal. Mach. Intelligence}, 23\penalty0
  (6):\penalty0 643--660, 2001.

\bibitem[Goh \& Ngo(2003)Goh and Ngo]{GoandNgo}
Goh, Alwyn and Ngo, David~C.L.
\newblock Computation of cryptographic keys from face biometrics.
\newblock In \emph{Communications and Multimedia Security. Advanced Techniques
  for Network and Data Protection}, volume 2828 of \emph{Lecture Notes in
  Computer Science}, pp.\  1--13. Springer Berlin Heidelberg, 2003.

\bibitem[He \& Niyogi(2004)He and Niyogi]{lpp}
He, X. and Niyogi, P.
\newblock Locality preserving projections.
\newblock \emph{Proc. of the NIPS, Advances in Neural Information Processing
  Systems. Vancouver: MIT Press}, 103, 2004.

\bibitem[He et~al.(2005)He, Cai, Yan, and Zhang]{npe}
He, Xiaofei, Cai, Deng, Yan, Shuicheng, and Zhang, Hong-Jiang.
\newblock Neighborhood preserving embedding.
\newblock In \emph{Computer Vision, 2005. ICCV 2005. Tenth IEEE International
  Conference on}, volume~2, pp.\  1208--1213 Vol. 2, Oct 2005.

\bibitem[Hegde et~al.(2007)Hegde, Wakin, and Baraniuk]{manifold2}
Hegde, Chinmay, Wakin, Michael~B., and Baraniuk, Richard~G.
\newblock Random projections for manifold learning.
\newblock In Platt, John~C., Koller, Daphne, Singer, Yoram, and Roweis, Sam~T.
  (eds.), \emph{NIPS}. Curran Associates, Inc., 2007.

\bibitem[Li et~al.(2006)Li, Hastie, and Church]{very_sparse}
Li, Ping, Hastie, Trevor~J., and Church, Kenneth~W.
\newblock Very sparse random projections.
\newblock In \emph{Proceedings of the 12th ACM SIGKDD international conference
  on Knowledge discovery and data mining}, KDD '06, pp.\  287--296, New York,
  NY, USA, 2006. ACM.

\bibitem[Roweis \& Saul(2000)Roweis and Saul]{lle}
Roweis, Sam~T. and Saul, Lawrence~K.
\newblock Nonlinear dimensionality reduction by locally linear embedding.
\newblock \emph{Science}, 290:\penalty0 2323--2326, December 2000.

\bibitem[Sarl{\'o}s(2006)]{dlogd}
Sarl{\'o}s, Tam{\'a}s.
\newblock Improved approximation algorithms for large matrices via random
  projections.
\newblock In \emph{FOCS}, pp.\  143--152. IEEE Computer Society, 2006.

\bibitem[Shakhnarovich \& Moghaddam(2004)Shakhnarovich and
  Moghaddam]{face_subspace}
Shakhnarovich, Gregory and Moghaddam, Baback.
\newblock Face recognition in subspaces.
\newblock In \emph{In: S.Z. LI, A.K. Jain, Handbook of Face Recognition}, pp.\
  141--168. Springer, 2004.

\bibitem[Shi et~al.(2012)Shi, Shen, Hill, and van~den Hengel]{classification}
Shi, Qinfeng, Shen, Chunhua, Hill, Rhys, and van~den Hengel, Anton.
\newblock Is margin preserved after random projection?
\newblock \emph{CoRR}, abs/1206.4651, 2012.

\bibitem[Sim et~al.(2002)Sim, Baker, and Bsat]{pie}
Sim, Terence, Baker, Simon, and Bsat, Maan.
\newblock The cmu pose, illumination, and expression (pie) database.
\newblock In \emph{Automatic Face and Gesture Recognition, 2002. Proceedings.
  Fifth IEEE International Conference on}, pp.\  46--51. IEEE, 2002.

\bibitem[Teoh et~al.(2006)Teoh, Goh, and Ngo]{rmq}
Teoh, A.B.J., Goh, A., and Ngo, D.C.L.
\newblock Random multispace quantization as an analytic mechanism for
  biohashing of biometric and random identity inputs.
\newblock \emph{IEEE Trans. Pattern Anal. Mach. Intelligence}, 28\penalty0
  (12):\penalty0 1892--1901, Dec 2006.

\bibitem[Vempala(2004)]{jl}
Vempala, S.
\newblock \emph{The Random Projection Method}.
\newblock Dimacs Series in Discrete Mathematics and Theoretical Computer
  Science, 2004.

\bibitem[Wright et~al.(2009)Wright, Yang, Ganesh, Sastry, and Ma]{src}
Wright, J., Yang, A.Y., Ganesh, A., Sastry, S.S., and Ma, Yi.
\newblock Robust face recognition via sparse representation.
\newblock \emph{IEEEE TPAMI}, 31\penalty0 (2):\penalty0 210 --227, Feb. 2009.

\end{thebibliography}
